\documentclass[sigconf]{acmart}
\usepackage{multirow}
\AtBeginDocument{%
  }

\DeclareMathOperator*{\argmax}{arg\,max}

\definecolor{darkgreen}{RGB}{0,100,0}   
\definecolor{darkred}{RGB}{139,0,0}   

\newcommand{\cmark}{\textcolor{darkgreen}{\scalebox{1.3}{\checkmark}}}  
\newcommand{\xmark}{\textcolor{darkred}{\textbf{\large $\times$}}}

\begin{document}

\title{PINE: Pipeline for Important Node Exploration in Attributed Networks}






\author{Elizaveta Kovtun}
\affiliation{
  \institution{Sber AI Lab, Skoltech}
  \country{Russia}
}

\author{Maksim Makarenko}
\affiliation{
  \institution{Sber AI Lab}
  \country{Russia}
}

\author{Natalia Semenova}
\affiliation{
  \institution{Sber AI, AIRI}
  \country{Russia}
}

\author{Alexey Zaytsev}
\affiliation{
  \institution{Skoltech}
  \country{Russia}
}

\author{Semen Budennyy}
\affiliation{
  \institution{Sber AI, AIRI}
  \country{Russia}
}

\begin{abstract}


A graph with semantically attributed nodes are a common data structure in a wide range of domains. It could be interlinked web data or citation networks of scientific publications. The essential problem for such a data type is to determine nodes that carry greater importance than all the others, a task that markedly enhances system monitoring and management. Traditional methods to identify important nodes in networks introduce centrality measures, such as node degree or more complex PageRank. However, they consider only the network structure, neglecting the rich node attributes. Recent methods adopt neural networks capable of handling node features, but they require supervision. This work addresses the identified gap--the absence of approaches that are both unsupervised and attribute-aware--by introducing a Pipeline for Important Node Exploration (PINE). At the core of the proposed framework is an attention-based graph model that incorporates node semantic features in the learning process of identifying the structural graph properties. The PINE's node importance scores leverage the obtained attention distribution. We demonstrate the superior performance of the proposed PINE method on various homogeneous and heterogeneous attributed networks. As an industry-implemented system, PINE tackles the real-world challenge of unsupervised identification of key entities within large-scale enterprise graphs.




\end{abstract}

\ccsdesc[500]{Networks~Network structure}
\ccsdesc[300]{Computing methodologies~Unsupervised learning}

\keywords{node importance estimation; attention mechanism; attributed networks}


\maketitle


\begin{figure*}[ht!]
\centering
\includegraphics[width=0.9\linewidth]{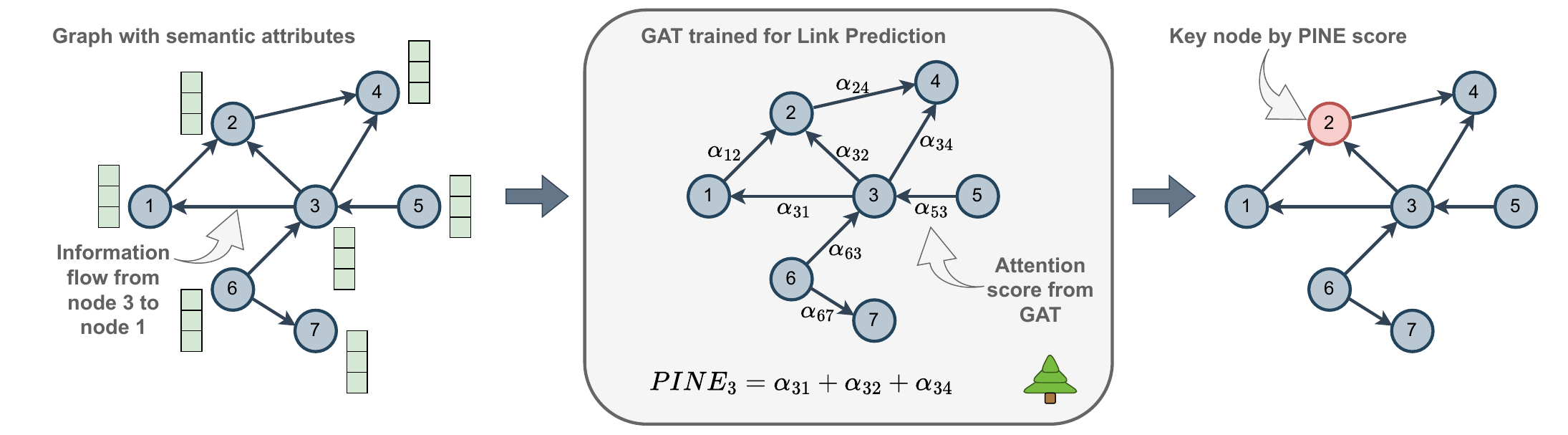}
\caption{PINE scheme. PINE is an approach for unsupervised identification of important nodes in directed networks that considers the node-level attributes. The importance score of a node is evaluated from the attention scores of GAT model trained with Link Prediction task. A direction of an edge from a node $v_i$ to a node $v_j$ means an information flow from $v_i$ to $v_j$. On the example of node 3, PINE score is calculated as a sum of attention weights between node 3 and nodes 1, 2, and 4. These attention weights reflect the extent of the usefulness of node 3 as an information provider to its neighbors. }
\label{fig:scheme_pine}
\end{figure*}


\begin{table}
\caption{Comparison of PINE and baselines with respect to awareness of topology and semantic information over network, applicability to knowledge graphs (KG), and need in ground truth markup within node importance estimation problem.}
\label{tab:figure_of_merit}
\begin{tabular}{ccccc}
\toprule
Setting & \shortstack{PageRank \\ \cite{page1999pagerank}} 
        & \shortstack{GENI \\ \cite{park2019estimating}} 
        & \shortstack{EASING \\ \cite{chen2025semi}} 
        & \shortstack{\textbf{PINE} \\ (ours)} \\
\midrule
Topology-aware & $\cmark$ & $\cmark$ & $\cmark$ & $\cmark$ \\
Semantic-aware & $\xmark$ & $\xmark$ & $\cmark$ & $\cmark$ \\
Applicability to KG & $\xmark$ & $\cmark$ & $\cmark$ & $\cmark$ \\
No need in markup & $\cmark$ & $\xmark$ & $\xmark$ & $\cmark$ \\
\bottomrule
\end{tabular}
\end{table}

\section{Introduction} 

Graphs offer a natural and flexible way to represent data across diverse domains. Depending on the context, nodes and edges may represent people and friendships in social networks \cite{Kahn2024}, stations and routes in transportation systems \cite{Leite2024}, or proteins and their interactions in biological networks \cite{Zhang2023}. As real-world graphs often consist of millions of nodes and edges with intricate topologies, performing fine-grained analysis can be computationally intensive, frequently requiring time and memory that scale superlinearly with graph size\cite{williams2010subcubic, newman2004finding}. A fundamental task in this context is the identification of key or influential nodes—those that play a pivotal role in maintaining the functionality, resilience, or information flow within the network \cite{kitsak2010identification, albert2000error}. This challenge arises in numerous web scenarios: detecting viral spreaders in social media~\cite{Drolsbach2023}; identifying authoritative sources in web search~\cite{yan2011discovering}; discovering critical users or items in recommendation systems~\cite{Tsugawa_2023}; and monitoring influence pathways in digital marketing~\cite{Shao2018}.


Existing methods for identifying important nodes fall into two broad categories: non-trainable and learning-based approaches. Non-trainable methods rely on predefined structural metrics such as degree \cite{bamakan2019opinion}, Closeness and Betweenness Centrality \cite{bloch2023centrality}, PageRank \cite{yan2011discovering}, Katz \cite{zhan2017identification}, and K-shell decomposition \cite{kitsak2010identification}. These methods estimate importance based solely on graph topology, entirely ignoring node-level semantic attributes. While such measures are interpretable and computationally attractive for smaller graphs, they capture only limited facets of influence and do not generalize well across domains or data types. Extensions such as hybrid metrics \cite{zhang2015identifying, ishfaq2022identifying} or improved formulations \cite{maji2020influential, ullah2021identification} provide incremental gains but remain confined to this structurally constrained paradigm.

In contrast, learning-based approaches, particularly those utilizing Graph Neural Networks (GNNs) \cite{bhattacharya2023detecting, zhang2022new, kou2023identify}, incorporate both graph structure and node features, underscoring the importance of vertex attributes. However, their industrial deployment introduces a major challenge: the need for ground truth importance labels. In most real-world scenarios, such labels are not available, necessitating the use of simulation-based methods for target generation. For example, node importance may be inferred from dynamic processes such as epidemic spreading (e.g., the SIR model \cite{yu2020identifying}) or computed from traditional metrics on a small subset of the data \cite{maurya2021graph}. Yet, simulation outcomes depend heavily on hyperparameter choices and reflect specific operational assumptions, limiting their generalizability and introducing inductive bias. This highlights a critical gap: the lack of an unsupervised framework that can assess node criticality by jointly leveraging \emph{both} structural and semantic information. In this study, we bridge this gap by proposing PINE, an unsupervised learning-based framework for identifying important nodes through joint operation over graph structure and node-level semantics. Table \ref{tab:figure_of_merit} highlights the specific problems addressed by PINE in comparison with existing methods.

Furthermore, the PINE framework tackles a concrete industrial challenge: unsupervised identification of key entities in large-scale enterprise graphs, encountered in diverse applied problems. We first deploy it for patent identification and technological influence tracing, where nodes correspond to patents and edges capture citations. The same pipeline is then adapted to a banking setting to map relationships among companies and counterparties, supporting a strategic decision-making. In production, PINE operates on an unlabeled heterogeneous graph comprising about 5 million business entities and 50 million counterparties linked by over 300 million edges that represent ownership, management, transaction, and textual relations. The system has been integrated into the bank’s internal analytics platform and is now used by multiple business departments to obtain influence scores for enterprise clients and counterparties across various banking scenarios.

We summarize the main contributions of this work as follows:
\begin{itemize}
    \item We introduce PINE (Pipeline for Important Node Exploration), an unsupervised framework for identifying influential nodes in attributed graphs. PINE jointly leverages graph structure and node-level semantic information and defines node importance through attention weights learned by a Graph Attention Network (GAT).

    \item We perform a comprehensive evaluation of PINE against a broad range of supervised and unsupervised baselines on public homogeneous and heterogeneous networks. The results on open-source data show that PINE achieves state-of-the-art performance for unsupervised methods, required for production deployment in banking and enterprise analytics.

    \item We validate PINE on two industrial applications: tracing technological influence in patent networks and analyzing enterprise-scale business graphs in the banking domain.

    \item We construct and release the Patent Influence Dataset to support validation and benchmarking of influence-detection methods in industrial settings. The dataset is derived from over 12,000 patents collected from Google Patents. Partial expert annotations are provided.
    
    
    
\end{itemize}

The code for PINE and Patent Influence Dataset are available at \url{https://github.com/E-Kovtun/node-importance-for-attributed-networks}.

\section{Related Work}
The task of identifying critical nodes is represented extensively in the literature. Most existing works focus on the application or elaboration of various graph measures. We treat such approaches as \textit{non-trainable}. Another school of thought is the use of machine learning and deep learning models as a tool for determining node importance scores. We denote this group of methods as \textit{learning-based} ones. 

\paragraph{\textbf{Non-trainable methods.}}

A well-proven way to find key nodes in the graph is to estimate the node-level centrality measures \cite{bloch2023centrality}. The values of such measures reflect node importance within the structure, but from different viewpoints. In the simplest case, one can calculate degree-based centrality measures \cite{bamakan2019opinion, chen2012identifying, zhao2017evaluating}. Despite the computational simplicity, they are not used universally because the field of action is too localized. To evaluate node significance based on its neighborhood and to capture the global network context, there are two widely used measures: closeness centrality and betweenness centrality, discussed in \cite{bloch2023centrality}. 

In a separate group, there are measures based on the idea that the significance of a node is greater if it is connected to essential nodes. The primary measures related to this group are eigenvector centrality \cite{spizzirri2011justification} and its variants \cite{gomez2019centrality}, such as PageRank \cite{yan2011discovering}, Katz \cite{zhan2017identification}, and HITS \cite{najork2007hits}. Another measure leveraged for identifying the graph's vital nodes is K-shell \cite{kitsak2010identification}. 
The authors of \cite{wang2020identifying} suggest a hybrid measure comprised of local node information entropy and global K-shell, enabling the selection of influential nodes not only from the higher shell. Additionally, some works focus on measuring sophisticated blending \cite{zhang2015identifying, ishfaq2022identifying} or enhancing the existing ones \cite{maji2020influential, ullah2021identification}. 

Despite the great diversity of the developed measures, they are confined to the treatment of the structural traits of the networks. These limitations motivate the development of learning-based approaches that can integrate structural and semantic information within a unified framework.

\paragraph{\textbf{Learning-based methods.}}

Beyond hand-crafted centrality measures, recent studies have approached influential-node identification as a supervised learning problem. Early work applied traditional machine-learning models trained on structural features. For example, \cite{zhao2020machine, rezaei2023machine} evaluate classical algorithms to predict node influence scores generated by the SIR spreading model. 

Modeling node interactions is crucial for identifying influential nodes, a capability that GNNs provide natively and that underlies the transition from conventional ML to graph-based neural models. 
The study \cite{munikoti2022scalable} regards the nodes with the greatest criticality as the ones whose removal drastically decreases the graph robustness. For the identification of such nodes, the GNN-based model is trained on a small representative subset of nodes, for which targets are calculated. The GAT-based architecture is proposed in \cite{liu2022nie} for mapping nodes' static features to their dynamic impacts, calculated with the cascading failure model, while \cite{yu2020identifying} trains the convolutional neural network to predict SIR-generated labels. The group of works \cite{bhattacharya2023detecting, zhang2022new, kou2023identify, wu2024graph, xiong2024vital} focuses on the construction of node descriptions that can be effectively used to train deep learning models. Some works adopt GNNs to learn the approximation of the centrality measures \cite{fan2019learning, maurya2021graph} to avoid the computationally expensive calculation of their exact values in the case of large graphs. 

The task of assessing node importance is also urgent for heterogeneous networks. In such graphs, the situation becomes more complex due to the presence of multiple types of edges and nodes, as well as their associated semantic attributes. Several studies focus on training graph neural networks to solve node importance estimation task \cite{park2019estimating, park2020multiimport, huang2021representation, huang2022estimating, chen2024deep, chen2025semi}. The authors of \cite{ma2025node} introduce virtual nodes and contrastive learning to enhance the expressiveness of the node embeddings for importance estimation.

Despite the versatility and generalization ability of learning-based approaches, their major limitation lies in the need for node-level ground truth labels. Such supervision is rarely available in industrial settings due to data privacy, annotation cost, and scale constraints. 


\section{Methodology}
This section presents the data at the core of our analysis and provides the formal definition of the important node identification task. Then, we discuss the propagation models that are critical elements for the quality assessment of the identified essential nodes. Further, we describe the proposed PINE approach that allows for the unsupervised estimation of node importance scores. Finally, we consider the performance metrics applicable to the considered task.

\subsection{Problem statement}
Let $G=(V, E)$ be directed homogeneous graph, where $V=\{v_1, \dots, v_N\}$ is the set of nodes and $E=\{e_1, \dots, e_M\}$ is the set of edges. $N = |V|$ and $M = |E|$ are the number of nodes and edges, respectively. 
In this work, the direction of an edge from node $v_i$ to $v_j$ means that information on $v_i$ is transferred to node $v_j$. Each node $v_i, 1 \leq i \leq N$, is attributed with the feature vector $\mathbf{x}_i \in \mathbb{R}^{d}$. Let $\mathbf{X} \in \mathbb{R}^{N \times d}$ denote the feature matrix with information on all nodes. We consider the important node identification problem from the influence maximization (IM) perspective \cite{li2018influence}. In solving the IM problem, one aims to find a group of vital nodes that causes the maximum information spread in a network, whether an information dissemination process starts from them. Formally, let $\phi(\cdot)$ denote the influence function associated with the number of activated nodes as a result of an information diffusion process. In the IM task, the goal is to find a subset $S^* \subseteq V$ of $K$ nodes, where $K < N$, such that it maximizes the influence over the graph $G$:

\begin{equation} 
S^* = \argmax_{S \subseteq V, |S| = K} \phi(S).
\end{equation}

\subsection{Propagation models}
In this work, we refer to the process of information propagation over the network as information diffusion or dissemination process. The models that define the rules of information diffusion are denoted as propagation models. Regarding the directed and attributed graphs, propagation models should respect the predetermined edge directions and consider node attributes to reflect a realistic picture. For this reason, we opt for Linear Threshold and Independent Cascade models upgraded with the attribute interaction between nodes, referred to as $\mathrm{LT+}$ and $\mathrm{IC+}$. In addition, we include considerations of a widespread SIR propagation model despite disregarding node-level information in a default setting.

\paragraph{\textbf{$\mathrm{LT+}$ propagation model.}}

The propagation model $\mathrm{LT+}$ is proposed in \cite{wang2022identifying}. It represents an upgraded version of the classical $\mathrm{LT}$ model \cite{chen2010scalable, kempe2003maximizing}. Within $\mathrm{LT}$ model, each node of a graph can be in an active or inactive mode. For the case of a directed graph, only a group of in-neighbors of node $v_i$ denoted as $\mathcal{N}_{in}(v_i)$ exerts a direct influence on it. Let $I(v_j, v_i)$ denote a measure of influence from node $v_j$ to node $v_i$. The aggregated influence from all in-neighbors of node $v_i$ must be no greater than one: $\sum_{j \in \mathcal{N}_{in}(v_i)} I(v_j, v_i) \leq 1$. Each node $v_i$ is assigned with an activation threshold $\theta_i \in [0, 1]$. A node $v_i$ can switch from an inactivate mode to an active one if the combined influence exceed the predefined threshold: $\sum_{j \in \mathcal{N}_{in}(v_i)} I(v_j, v_i) \geq \theta_i$. The diffusion continues until no additional nodes are eligible for activation. 

$\mathrm{LT+}$ propagation model allows concurrent consideration of the structural and the semantic interplay between nodes. The structural influence of node $v_j \in \mathcal{N}_{in}(v_i) $ on node $v_i$ is defined as:

\begin{equation}
I^{\mathrm{struct}}(v_j, v_i) = \frac{1}{\mathrm{in\text{-}degree}(v_i)}, 
\end{equation}
where $\mathrm{in\text{-}degree}(v_i)$ is a number of in-neighbors of node $v_i$.

Node attributes can significantly affect the strength of impact between nodes. Intuitively, a greater similarity between node attributes points to a potential of the stronger interaction. To evaluate attribute similarity, we adopt a cosine measure: $\mathrm{sim}(v_j, v_i) = \frac{\mathbf{x}_j \mathbf{x}_i}{||\mathbf{x}_j|| \cdot ||\mathbf{x}_i||} $. The semantic influence of node $v_j$ on node $v_i$ is given by:

\begin{equation}
I^{\mathrm{sem}}(v_j, v_i) = \frac{\exp(\mathrm{sim}(v_j, v_i))}{\sum_{v_l \in \mathcal{N}_{in}(v_i)}\exp(\mathrm{sim}(v_l, v_i))}
\end{equation}
Notably, there is an edge-softmax normalization by destination nodes for balancing purposes. The overall incoming influence for a node $v_i$ from a node $v_j$ is calculated as:
\begin{equation} \label{prob} 
I(v_j, v_i) = \alpha_1 \cdot I^{\mathrm{struct}}(v_j, v_i) + \alpha_2 \cdot I^{\mathrm{sem}}(v_j, v_i),
\end{equation}
where parameters $\alpha_1$ and $\alpha_2$ ($\alpha_1 + \alpha_2 = 1$) control a contribution of structural and semantic properties of the network to the information diffusion process, respectively. This joint consideration of structural proximity and semantic similarity reflects real-world dynamics.

\paragraph{\textbf{$\mathrm{IC+}$ propagation model.}}
$\mathrm{IC}$ is a popular propagation model \cite{kempe2003maximizing}. Analogously to $\mathrm{LT}$ setup, nodes can exist in either active or inactive states. The primary mechanism of $\mathrm{IC}$ consists of assigning each edge the probability of activating a node under the influence of another one. We generalize a basic version of $\mathrm{IC}$ to $\mathrm{IC+}$, as is done with $\mathrm{LT}$. In $\mathrm{IC+}$, a chance for $v_j$ to activate initially inactive node $v_i$ is defined by equation \ref{prob}. Such a probability definition intends to inject semantic relationships between nodes in a standard $\mathrm{IC}$ propagation model. Other variants of a semantic enrichment are given in \cite{hong2019seeds, li2024ra}

\paragraph{\textbf{$\mathrm{SIR}$ propagation model.}}
$\mathrm{SIR}$ \cite{kermack1927contribution} is a well-established propagation model, stemming from the studies on infectious disease spreading through a population. 
In the original version, SIR does not account for node attributes. We include it for a fair comparison with other works in which SIR is a primary propagation model. 

\begin{table*}[ht!]
\caption{The resulting influence spread values when using different methods to select influential starting nodes. The top 10\% of nodes for each dataset are selected as seed nodes. The performance is reported within respect to three diffusion models $\mathrm{LT+}$, $\mathrm{IC+}$, and $\mathrm{SIR}$. The best values are \textbf{highlighted}, the second best are \underline{underscored}.}
\label{tab:metric_res}
\begin{tabular}{cccccccccccc}
\toprule
Data & & Degree & Out-degree & Weighted & Relative & PageRank & VoteRank & Katz & BC & EnRenew & \textbf{PINE} (ours)\\
\hline
\multirow{3}{*}{Cora}
& $\mathrm{LT+}$ & 0.596 & 0.642 & 0.631 & 0.647 & 0.620 & 0.651 & 0.653 & 0.537 & \textbf{0.733} & \underline{0.688} \\
& $\mathrm{IC+}$ & 0.501 & 0.527 & 0.520 & 0.531 & 0.513 & 0.540 & 0.534 & 0.458 & \textbf{0.569} & \underline{0.567} \\
& $\mathrm{SIR}$ & 0.278 & 0.298 & 0.289 & 0.300 & 0.278 & 0.296 & 0.300 & 0.242 & \underline{0.301} & \textbf{0.304}\\
\hline
\multirow{3}{*}{CiteSeer}
& $\mathrm{LT+}$ & 0.329 & 0.442 & 0.433 & 0.454 & 0.481 & 0.438 & 0.448 & 0.327 & \underline{0.494} & \textbf{0.511} \\
& $\mathrm{IC+}$ & 0.302 & 0.392 & 0.384 & 0.402 & \underline{0.430} & 0.397 & 0.395 & 0.299 & 0.422 & \textbf{0.457} \\
& $\mathrm{SIR}$ & 0.169 & 0.216 & 0.211 & \underline{0.218} & 0.215 & 0.203 & \underline{0.218} & 0.169 & \underline{0.218} & \textbf{0.228} \\
\hline
\multirow{3}{*}{PubMed}
& $\mathrm{LT+}$ & 0.690 & 0.797 & 0.779 & 0.794 & 0.853 & 0.842 & 0.786 & 0.718 & \textbf{0.901} & 0.\underline{869} \\
& $\mathrm{IC+}$ & 0.626 & 0.713 & 0.696 & 0.709 & 0.759 & 0.754 & 0.702 & 0.654 & \underline{0.765} & \textbf{0.776} \\
& $\mathrm{SIR}$ & 0.298 & 0.343 & 0.336 & 0.342 & \underline{0.356} & 0.352 & 0.340 & 0.301 & \underline{0.356} & \textbf{0.363} \\
\hline
\multirow{3}{*}{Wiki-CS}
& $\mathrm{LT+}$ & 0.905 & 0.914 & 0.915 & 0.916 & 0.934 & \underline{0.941} & - & 0.940 & 0.906 & \textbf{0.946} \\
& $\mathrm{IC+}$ & 0.589 & 0.598 & 0.600 & 0.600 & 0.622 & \underline{0.632} & - & 0.630 & 0.582 & \textbf{0.638} \\
& $\mathrm{SIR}$ & 0.743 & 0.746 & 0.747 & 0.744 & 0.749 & 0.748 & - & \underline{0.750} & 0.745 & 0.\textbf{753} \\
\hline
\multirow{3}{*}{HEP-TH}
& $\mathrm{LT+}$ & 0.409 & 0.565 & 0.565 & 0.566 & 0.726 & 0.615 & - & 0.548 & \textbf{0.762} & \underline{0.740} \\
& $\mathrm{IC+}$ & 0.323 & 0.424 & 0.425 & 0.426 & \underline{0.527} & 0.460 & - & 0.418 & 0.510 & \textbf{0.541} \\
& $\mathrm{SIR}$ & 0.350 & 0.441 & 0.439 & 0.440 & 0.514 & 0.464 & - & 0.431 & \textbf{0.540} & \underline{0.526} \\
\hline
\multirow{3}{*}{ogbn-Arxiv}
& $\mathrm{LT+}$ & 0.630 & 0.736 & 0.734 & 0.734 & 0.833 & 0.787 & - & - & \textbf{0.863} & \underline{0.843} \\
& $\mathrm{IC+}$ & 0.471 & 0.549 & 0.547 & 0.548 & 0.611 & 0.586 & - & - & \underline{0.561} & \textbf{0.627} \\
& $\mathrm{SIR}$ & 0.441 & 0.502 & 0.502 & 0.502 & 0.554 & 0.530 & - & - & \textbf{0.560} & \underline{0.558} \\
\hline
DBLP & $\mathrm{SIR}_{100}$ & 0.409 & \underline{0.463} & \underline{0.462} & \underline{0.463} & - & - & - & - & - & \textbf{0.491} \\
\hline
\end{tabular}
\end{table*}

\begin{figure}
\centering
\includegraphics[width=\linewidth]{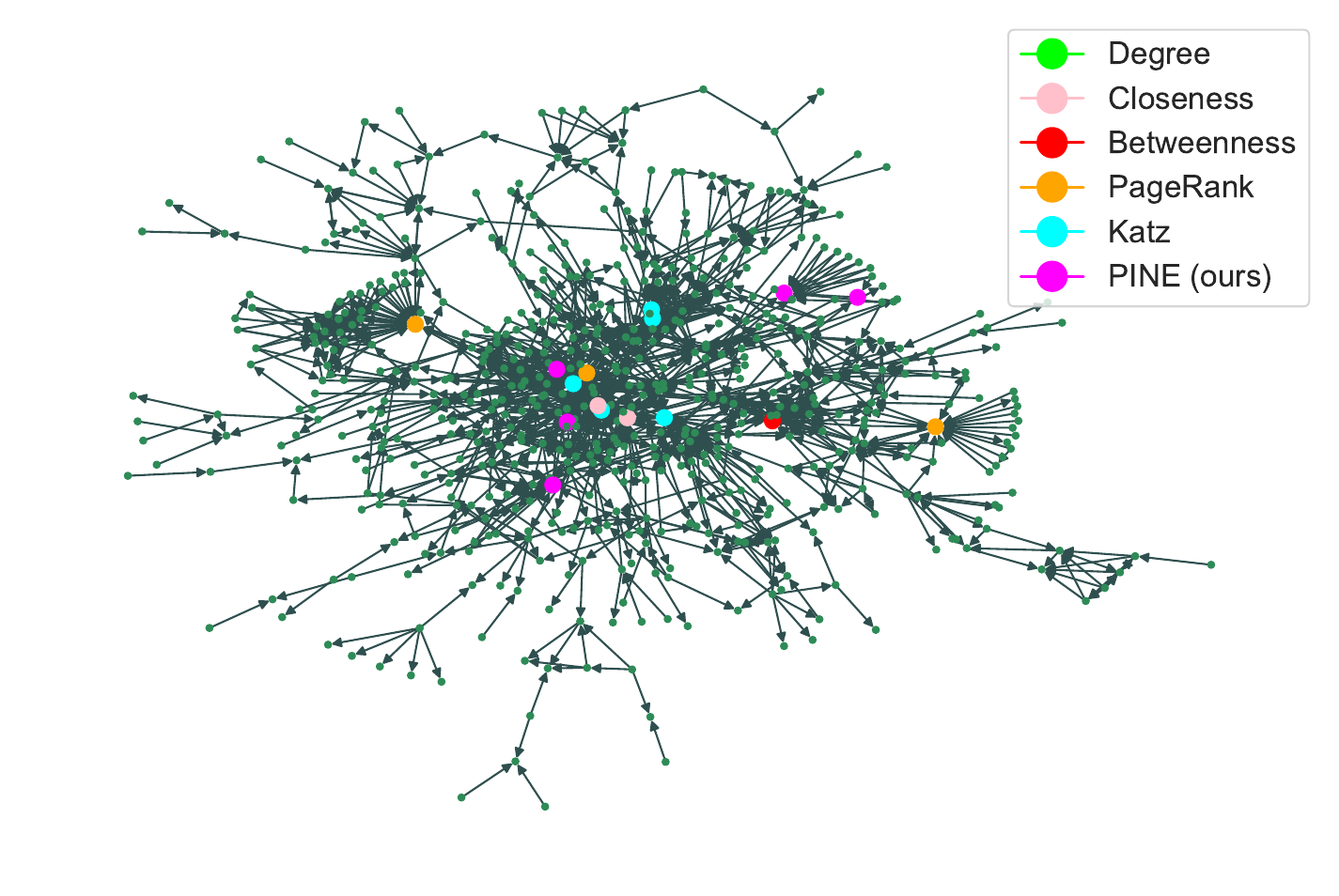}
\caption{Patent citation graph for physical photoresist model. Each approach highlights a top-5 important nodes from its perspective.}
\label{fig:graph_photoresist}
\end{figure}

\subsection{PINE approach}
We propose a PINE framework that allows for deriving nodes' importance scores jointly from the graph structure and the node attributes in an unsupervised regime. At the pipeline's core is the training of a graph attention model (GAT) to solve the Link Prediction (LP) task. During training on this task, attention weights within GAT are distributed to enable effective node representation updates for inferring edge presence. From a specific perspective, the resulting attention weights assigned to edges can be a measure of each node's usefulness as an information provider for its neighbors. The fundamental idea behind PINE is that important nodes act as a critical information source for most adjacent nodes. This claim will be proved empirically in Section \ref{exps_section}. As output, PINE binds an importance score to each node in the network, derived by accumulating its utility to surrounding nodes. We treat PINE as an unsupervised approach for solving the node importance estimation task, since it does not utilize any ground truth importance markup. We further describe each component of PINE in detail. 

\subsubsection{Attention mechanism in PINE}
The key phase of PINE framework is to train GAT \cite{velivckovic2017graph} to solve the LP task. The motivation for the incorporation of the GAT architecture in the pipeline is twofold. Firstly, an attention mechanism of GAT facilitates estimating the degree of node interconnections in a learnable and data-driven way, without relying on any heuristic assumption. Secondly, the logic of GAT operation is transparent and interpretable compared to more sophisticated attention-based models on graphs, thus making adaptation and application of its attention mechanism straightforward for subsequent use. The selection of the LP problem solved within PINE is also deliberate. The LP task involves determining the existence of an edge between two nodes. Hence, it can be formulated for any graph, unlike the node classification task, which requires the presence of annotated nodes. Moreover, the solution of the LP problem inherently necessitates deep exploitation of structural and semantic graph properties, thereby producing attention weights enriched with both types of information.  

GAT architecture consists of $L$ layers. We describe the process within a specific $l$-th layer, $l = 1, \dots, L$, omitting the layer index for brevity. A rule for updating the representation of node $v_i$ in a GAT layer is the following:
\begin{equation}
\widehat{\mathbf{h}}_{i} = \sum_{j \in \mathcal{N}_{in}(v_i)}\alpha_{ji}\mathbf{U}\mathbf{h}_{j},
\end{equation}
where $\mathbf{h}_{j} \in \mathbb{R}^{d}$ is a representation of node $v_j$ at the input of the current layer. Prior to input into the model, nodes are described by $\mathbf{X} \in \mathbb{R}^{N \times d}$ matrix with semantic embeddings for each node. $\mathcal{N}_{in}(v_i)$ is in-neighbors of node $v_i$, $\mathbf{U} \in \mathbb{R}^{\hat{d} \times d}$ is a learnable weight matrix, and $\alpha_{ji}$ is an attention weight. The edge directed from node $v_j \in \mathcal{N}_{in}(v_i)$ to $v_i$ means that there is an information flow from $v_j$ to $v_i$. Thus, the updated embedding $\widehat{\mathbf{h}}_{i} \in \mathbb{R}^{\widehat{d}}$ of node $v_i$ comprises information from all its in-neighbors $v_j \in \mathcal{N}_{in}(v_i)$, scaled by attention weights $\alpha_{ji}$ to control the extent of the information transmission. We intentionally omit the representation $\mathbf{h}_i$ of node $v_i$ itself in the update rule since our goal is for the model to focus on information exchange between adjacent nodes. Additionally, we adopt a single-head attention for the sake of preserving the straightforward usability of layer-wise attention distributions. The attention weight for an edge directed from node $v_j$ to $v_i$ is calculated as follows:
\begin{equation}
\alpha_{ji} = \frac{\exp(w_{ji})}{\sum_{j \in \mathcal{N}_{in}(v_i)} \exp(w_{ji})},
\end{equation}
where the normalization is performed over the edges incoming to the considered node $v_i$. The unnormalized coefficient for estimating the degree of influence from the node $v_j$ to the node $v_i$ is defined as:
\begin{equation}
w_{ji} = \mathrm{LeakyReLU}\left((\mathbf{U}\mathbf{h}_j, \mathbf{s}) + (\mathbf{U}\mathbf{h}_i, \mathbf{t})\right). 
\end{equation}
$(\cdot, \cdot)$ means the scalar product, carried out between the projected node representations and trainable matrices $\mathbf{s}, \mathbf{t} \in \mathbb{R}^{\widehat{d}}$. $\mathrm{LeakyReLU}$ is an activation function defined in \cite{maas2013rectifier}.

\subsubsection{Link prediction within PINE}
After application of an attention mechanism for updating node embeddings, we solve the LP task on the basis of the resulting representations. Let $\mathbf{h}_j^{out}$ and $\mathbf{h}_i^{out}$ denote the embeddings of nodes $v_j$ and $v_i$ after passing through all $L$ layers of GAT. The probability of an edge existing between nodes $v_j$ and $v_i$ is given by%
\begin{equation}
z_{ji} = \mathrm{\sigma}\left((\mathbf{h}^{out}_j, \mathbf{h}^{out}_i)\right), 
\end{equation}
where $\sigma$ is a sigmoid function. For model training within LP task, we use a binary cross-entropy loss defined as:
\begin{equation}
\mathcal{L}_{LP} = - \sum_{ji} \left(y_{ji} \log{z_{ji}} + (1 - y_{ji}) \log{(1-z_{ji})}\right), 
\end{equation}
where the summation is performed over the sampled pairs of nodes $v_j$ and $v_i$ with the existing ($y_{ji}=1$) and non-existing ($y_{ji}=0$) edges between them. The numbers of positive and negative edges are balanced in a one-to-one ratio in a loss function. 

\subsubsection{Calculation of PINE importance scores}
As a result of training the network on the link prediction task, there is a valuable distribution of topology- and semantic-enriched attention weights. As previously stated, the importance of a node within the PINE framework is associated with the significance of this node as an information supplier for its out-neighbors. In line with this argument, the PINE importance score of node $v_j$ is calculated as: 

\begin{equation}\label{PINE} \mathrm{PINE}_j = \sum_{i \in \mathcal{N}_{out}(v_j)} \alpha_{ji} \end{equation}

Thus, if neighbors $v_i \in \mathcal{N}_{out}(v_j)$ attend to the considered node $v_j$ with significant weights to acquire relevant information, it means that the node $v_j$ plays a crucial structural and semantic role within the network. Figure \ref{fig:scheme_pine} depicts the logic behind the PINE scores calculation procedure.

\subsubsection{Adaptation of PINE to heterogeneous graphs} \label{pine_heterogeneous}
In practical scenarios, the problem of identifying key nodes commonly arises in a heterogeneous graph. Assume a graph where all nodes are of the same type, but edges belong to different types $P=\{p_1, \dots, p_B\}$. A direct adaptation of PINE involves applying it to $B$ homogeneous subgraphs constituting the original heterogeneous network. Notably, not all subgraphs originating from the particular edge type provide a meaningful knowledge source for deriving importance scores. Conversely, their consideration may introduce noise into the overall importance scores. To deal with this problem, we need to preliminarily identify a set of valuable edge types $P^{*}$ to sum PINE scores over them:
\begin{equation} 
\mathrm{PINE}^{\mathrm{heterogeneous}}_j = \sum_{p_b \in P^*} \; \sum_{i \in \mathcal{N}_{out}(v_j)} \alpha^{(p_b)}_{ji},
\end{equation}
where $\alpha^{(p_b)}_{ji}$ is attention weights over subgraph with edge type $p_b$.

\subsection{Performance metrics}
We evaluate node-importance methods under two complementary settings: unsupervised and supervised.
The unsupervised setup's true set of key nodes is unknown since the influence maximization problem is NP-hard. Evaluation relies on simulating a diffusion process initiated from the top-$K$ predicted nodes. The resulting proportion of activated nodes defines the \textit{influence spread}
\begin{equation}
\sigma(S) = \frac{\phi(S)}{n},
\end{equation}
where $\phi(S)$ is the number of activated nodes averaged over multiple Monte Carlo runs and $n$ is the total number of nodes.
The supervised setup assumes available node-importance annotations. Predicted importance scores are compared against ground-truth values using ranking and correlation metrics: Normalized Discounted Cumulative Gain ($\mathrm{NDCG}$), Spearman’s rank correlation coefficient, and $\mathrm{Precision@k}$. Their definitions are given in Appendix \ref{supervised_metrics}.

\section{Experiments and Results} \label{exps_section}
In this section, we discuss the used datasets and demonstrate PINE performance compared to other established methods. Then, we introduce an industrial use-cases with PINE.

\subsection{Datasets}
A primary requirement for dataset selection in analyzing PINE performance is the presence of semantic features in nodes. In this work's context, semantics in a network refers to the ability of nodes to acquire knowledge from one another. In addition, nodes should inherently embody a property of importance to facilitate the study of key node identification. Intuitively, a significant entity carries essential information upon which the other nodes rely. Networks with scientific publications and web pages adequately meet these criteria. The connections in such networks are governed by citation relationships or the presence of hyperlinks that entail a knowledge transfer. 

As for homogeneous graphs with scientific papers in nodes, we take Cora, CiteSeer, PubMed, HEP-TH, ogbn-Arxiv, and DBLP datasets. They constitute graphs of various scales and are composed of publications from different fields. Wiki-CS dataset pertains to web data as it is composed of the Wikipedia articles. To illustrate the generalization capability of PINE, we also consider a heterogeneous network called FB15K with 1,345 edge types. FB15K is also related to the web and is accompanied by external annotation of node importance, which corresponds to the number of webpage views. In the discussed networks, the semantic information of nodes, which originates from abstracts and titles of papers or content of webpages, is encoded into embeddings. The characteristics of networks, including numbers of nodes and edges, the dimension of semantic embeddings, are given in Table \ref{tab:graph_stats}. More detailed description of each dataset is provided in Appendix \ref{datasets_info}.

\paragraph{Patent Influence Dataset for industrial use case}

We introduce a new partially labeled patent citation dataset designed to evaluate PINE in industrial settings. Each patent is treated as a node, and citations form directed edges, capturing technological influence.

The dataset is built from Google Patents\footnote{\url{https://patents.google.com}} using four queries: etch simulation, photolithography simulation, physical photoresist model, and plasma discharge simulation, covering patents issued between 2000 and 2024. For each domain, we construct a separate citation graph (statistics in Table~\ref{tab:graph_stats}). A subset of patents is expert-annotated for relevance and technological significance, providing weak supervision for evaluation. 
Related works on patent analysis and the details on processing patent data are summarized in Appendices~\ref {patent_works} and \ref{patent_data}, respectively.

\begin{table}
\caption{The graphs of different scales are under consideration. They represent the citation networks of scientific papers or the interlinked web graphs. Emb size refers to the dimensionality of semantic embedding. associated with each node in a graph. FB15K dataset constitutes a heterogeneous graph. In the lower part, there are characteristics of Patent Influence Datasets used within industrial use-case.}
\label{tab:graph_stats}
\begin{tabular}{cccccc}
\toprule
Dataset & \# Nodes & \# Edges & Emb size & Scale \\
\midrule
Cora & 2,708 & 5,429 & 1433 & Small\\
CiteSeer & 3,312 & 4,715 & 3703 & Small \\
PubMed & 19,717 & 44,338 & 500 & Medium \\
Wiki-CS & 11,367 & 297,110 & 300 & Medium \\
Hep-TH & 12,468 & 92,377 & 768 & Medium \\
ogbn-Arxiv & 169,343 & 1,166,243 & 768 & Large \\
DBLP & 1,106,759 & 6,120,897 & 768 & Large \\
\midrule
FB15K & 14,951 & 592,213 & 768 & Medium \\
\midrule
\multicolumn{5}{l}{\textbf{Patent Influence Dataset}} \\
\cmidrule(r){1-5}
Etch & 984 & 8028 & 2304 & Small \\
Photolithography & 345 & 893 & 2304 & Small \\
Photoresist & 593 & 1407 & 2304 & Small \\
Plasma & 550 & 1949 & 2304 & Small \\
\bottomrule
\end{tabular}
\end{table}

\subsection{Baselines} \label{baselines}
We perform a comparison of PINE with two type of methods: non-trainable and learning-based ones. 

\paragraph{\textbf{Non-trainable methods}} 
We examine three classes of non-trainable baselines, including degree-based, iterative, and global centrality measures. 

PINE idea aligns most closely with \textit{degree-based measures}, since it leverages a local neighborhood over a few hops to derive node level of importance. Therefore, we consider different existing local and semi-local centrality measures as primary baselines to compare with.

\begin{itemize}
\item{\textbf{Node Degree.}} It counts a total number of in- and out-neighbors that the considered node has. 
\item{\textbf{Out-degree.}} This measure accounts only for out-neighbors.
In the context of this study, out-degree reflects the number of nodes to which the considered node transmits information.
\item{\textbf{Weighted out-degree.}} Let each edge be attributed with a cosine similarity of node attributes that it connects, then a weighted out-degree is defined as a sum of weights of its out-edges. 
\item{\textbf{Relative out-degree \cite{opsahl2010node}.}} This measure performs aggregation of node out-degree and its weighted version with a tuning parameter that controls a relative contribution of each component.
\end{itemize}

A popular way to discover influential nodes is to adopt \textit{iterative centrality measures}. They operate by repeatedly updating node scores through propagation of influence across the network until convergence. Their iterative mechanism enables capturing not only local node properties but also global structural information, including indirect relationships establishes by multi-step paths. Note that these methods inherently require careful calibration of parameters, like damping factors or attenuation coefficients, to ensure convergence and meaningful interpretation. We select several well-established approaches along with the recent ones for a subsequent performance comparison.

\begin{itemize}
\item{\textbf{PageRank (PR) \cite{page1999pagerank}.}} The basic idea behind PageRank, originally designed for the ranking of web pages, is to regard links coming from important pages as more significant than links from obscure pages. 

\item{\textbf{Katz Centrality \cite{katz1953new}.}} By the idea of Katz measure, the node is important if it is highly linked by other nodes or if links to it come from important nodes. 

\item{\textbf{VoteRank \cite{liu2021identifying}.}} VoteRank is a voting approach for identification of influential nodes \cite{zhang2016identifying}. 

\item{\textbf{EnRenew \cite{guo2020influential}.}} The EnRenew is an algorithm for identification of a set of influential nodes via information entropy measurements.
\end{itemize}

We also incorporate into our analysis \textit{global centrality measures}. They quantify node importance based on its position relative to the entire network topology. Although they provide insights on a global role of a node in the graph, their computation is computationally expensive and can become infeasible for large-scale graphs.

\begin{itemize}
\item{\textbf{Closeness Centrality (CC) \cite{freeman1979centrality, bloch2023centrality}}}. This measure builds upon the distances from the considered node to every other node in a graph. 

\item{\textbf{Betweenness Centrality (BC) \cite{freeman1977set, bloch2023centrality}}}. The values of BC point out the importance of a node as an intermediary in the information transmission between other nodes in the graph. 
\end{itemize}

\paragraph{\textbf{Learning-based methods}} 
With the development of graph neural networks, their application to the task of identifying important nodes gains increasing popularity. They are commonly employed for identifying node importance scores in knowledge graphs. 

\begin{itemize}
\item{\textbf{GENI \cite{park2019estimating}}}. GENI is a pioneering work for the problem of estimating node importance. GENI applies an attentive GNN, which distinguishes between various types of edges, and performs direct aggregation of important scores from neighbors internally.

\item{\textbf{RGTN \cite{huang2021representation}}}. Relational Graph Transformer Network (RGTN) approach uses separate encoders for semantic and structural node information. Co-attention module fuses representations of these types to produce node importance estimation.

\item{\textbf{EASING \cite{chen2025semi}}}. EASING is a semi-supervised node importance estimation framework that leverages unlabeled data to improve learning quality. EASING has Transformer-based architecture underneath for processing heterogeneous graphs. 

\end{itemize}

\subsection{Experimental results}
Before obtaining PINE importance scores, we need to train GAT model to solve Link Prediction task on graph under consideration. A high-quality training on LP task is a crucial stage of PINE to receive a reliable attention weight distribution. All technical details of this step are described in Appendix \ref{lp_training}. After that, we analyze the resulting attention distributions of trained model and derive PINE-based importance scores. Below, we compare key nodes discovered by PINE with ones produced by non-trainable and learning-based baselines.

\subsubsection{Comparison with non-trainable baselines}
To begin with, we compare PINE approach with non-trainable baselines discussed in Section \ref{baselines}. Each method associates nodes with some importance values. We select the top 10\% of nodes as starting nodes for launching information diffusion. Three propagation models, namely $\mathrm{LT+}$, $\mathrm{IC+}$, and SIR, are utilized in a dissemination process. 1000 Monte Carlo simulations of information diffusion are conducted for small-scale graphs, and 100 runs are done for medium- and large-scale networks. The methods are compared in Table \ref{tab:metric_res} by resulting influence spread values. Standard deviations are up to 0.003.

In most cases, PINE finds out more critical nodes. Iterative centrality measures EnRenew, PageRank, and VoteRank demonstrates solid performance on the majority of datasets. Notably, PINE scores include only 1-hop or 2-hops information on neighboring nodes, depending on the number of GAT layers. Still, it significantly outperforms degree-based measures, like Degree, Out-degree, Weighted, and Relative. These results suggest that during training on the LP task, the GAT model learns to discriminate the usefulness of information flows between neighbors effectively. Katz and Betweenness are not calculated for a part of medium- and large-scale networks because of prohibitively long execution time.

Due to their fast execution, PINE is compared with Degree, Out-degree, Weighted, and Relative methods on large-scale DBLP graphs. For each top 10\% nodes, we run an SIR propagation model with a limit of 100 iterations. The standard deviation in such experiments is equal to 0.001. PINE confirms its dominance even on graphs of this magnitude. The comparison of PINE computation time with other baselines are presented in Appendix \ref{computation_time}. Notably, PINE performs on par or faster than iterative-based and global measures. 

\subsubsection{Comparison with learning-based methods}
Here, we examine PINE effectiveness in the scope of heterogeneous graphs. We take FB15K dataset as a popular benchmark for node importance estimation task on heterogeneous networks. All nodes of FB15K are divided into train-valid-test sets in 1:1:1 ratio. Each set comprises 10\% of the overall node number with available markup. Three learning-based approaches, including GENI, RGTN, and EASING, are trained with a supervised signal from node importance ground truth. Validation set is used for early stopping. In this setting, we use heterogeneity-tailored version of PINE, discussed in Section \ref{pine_heterogeneous}. As FB15K consists of 1,345 of edge type, we leverage validation set to identify a type of edges that makes a meaningful contribution to the task key node detection problem. To do that, we launch PINE framework on top-100 subgraphs of a particular edge type, sorted by size. Subsequently, we select such edge types which exhibit positive $\mathrm{Spearman}$ correlation between type-related PINE scores and overall node ground truth. Besides, we calibrate PINE outputs by the values of nodes' out-degree computed across the whole network. A calibration of model outputs by degree centrality is a wide-spread technique, which, in particular, is utilized in GENI and EASING. Noteworthy, PINE does not utilize training node markup in any way. The comparison of PINE with supervised baselines is provided in Table \ref{tab:supervised_res}. PINE demonstrates strong results given the fact that it does not directly adopt ground truth annotation of nodes. Remarkably, PINE achieves the highest $\mathrm{Precision@100}$ score compared with all other methods, evidencing its strong capabilities in identifying sets of top-$k$ important nodes.

\begin{table}
\caption{A comparison of PINE with learning-based models on FB15K heterogeneous network. While GENI, RGTN, and EASING are trained on 10\% of nodes with annotation, PINE does not utilize any training supervision signal.}
\label{tab:supervised_res}
\begin{tabular}{ccccc}
\toprule
Method & $\mathrm{NDCG@100}$ & $\mathrm{Spearman}$ & $\mathrm{Precision@100}$  \\
\midrule
Out-degree & 0.835 $\pm$ 0.024 & 0.328 $\pm$ 0.020 & 0.272 $\pm$ 0.021 \\
\midrule
GENI & \textbf{0.948} $\pm$ 0.010 & \underline{0.460} $\pm$ 0.102 & \underline{0.290} $\pm$ 0.033 \\
RGTN & \textbf{0.947} $\pm$ 0.024 & \textbf{0.473} $\pm$ 0.189 & 0.204 $\pm$ 0.132 \\
EASING & 0.869 $\pm$ 0.059 & 0.423 $\pm$ 0.157 & \underline{0.292} $\pm$ 0.102 \\
\midrule
\textbf{PINE} (ours) & \underline{0.921} $\pm$ 0.013 & 0.402 $\pm$ 0.027 & \textbf{0.378} $\pm$ 0.019 \\
\bottomrule
\end{tabular}
\end{table}




\section{PINE industrial use-cases}

\paragraph{\textbf{Core patent identification.}}

\begin{table}
\caption{Expert ranking of the list of top-5 core patents for every technological domain obtained with different approaches. A smaller rank means the identified patents are of higher importance for the considered domain. The lowest rank (6) indicates an irrelevance of the provided patent group.}
\label{tab:pine_res}
\begin{tabular}{ccccccc}
\toprule
Key word & Degree & CC & BC & PR & Katz & \textbf{PINE}\\
\midrule
Etch & 4 & 3 & 4 & 2 & 5 & 1 \\
Photolithography & 4 & 2 & 3 & 5 & 6 & 1 \\
Photoresist & 2 & 4 & 5 & 6 & 1 & 3 \\
Plasma & 6 & 6 & 2 & 6 & 6 & 1 \\
\midrule
Mean rank & 4 & 3.75 & \underline{3.5} & 4.75 & 4.5 & \textbf{1.5} \\
\bottomrule
\end{tabular}
\end{table}

We compare PINE performance with baselines from \ref{baselines} on Patent Influence Datasets. The results are presented in Table \ref{tab:pine_res}. Each approach mentioned in Table \ref{tab:pine_res} calculates the importance score for all nodes. Then, these values are sorted and top-5 patents are taken. The lists of 5 patents of the highest importance are treated as the lists of core patents that are found with the mechanism of the particular method. The experts rank the outputs of every method from 1 to 6. Rank 1 means that the list mostly comprises the significant patents. Rank 6 indicates that the identified patents are irrelevant to domain or are of minor importance as a technological solution. PINE produces the list of patents, most of which are recognized as particularly important by experts.  Figure \ref{fig:graph_photoresist} demonstrates the positions of highlighted nodes by the baselines and PINE for the physical photoresist model.

\paragraph{\textbf{Key enterprise discovery in Bank.}}


Currently, PINE is applied for supporting strategic decision-making and business development. The system helps identify key enterprises and counterparties that play central roles in particular sectors or regions, allowing the bank to prioritize strategic clients, design targeted offers, and plan partnerships more effectively. PINE analyzes a heterogeneous graph of ownership, management, and transaction relations to uncover hidden interrelations. Besides, bank clients are described by a broad set of varied attributes. PINE accounts for them in its workflow, generating structure- and feature-informed importance scores. In this setting, PINE serves as a decision-support tool for relationship management, corporate lending, and strategic planning, guiding data-driven resource allocation across the enterprise client network.

\section{Conclusion}
In this work, we presented PINE (Pipeline for Important Node Exploration) — an unsupervised framework for identifying influential nodes in attributed networks by jointly leveraging structural connectivity and semantic node characteristics. The framework employs an attention-based graph model, where importance scores are derived directly from learned attention distributions, eliminating the need for external supervision or annotated data through the use of a link prediction pretext task.
The evaluation across public citation and industrial datasets shows that PINE consistently outperforms classical baselines such as PageRank, VoteRank, and degree-based measures, achieving up to 20\% higher influence spread across multiple propagation models. Despite being fully unsupervised, it also matches or surpasses supervised methods on heterogeneous benchmarks like FB15K. PINE has been validated in two real-world deployment scenarios: tracing technological influence in patent networks and mapping enterprise relationships in the banking domain.
To support further research, we also introduced the Patent Influence Dataset, a domain-grounded benchmark comprising over 12,000 patents and expert annotations across multiple technological domains. Overall, PINE represents a step toward industrial adoption of unsupervised neural methods 
as a practical tool for uncovering influential entities.

\bibliographystyle{ACM-Reference-Format}
\bibliography{sample-base}


\appendix

\section{Datasets} \label{datasets_info}
\begin{itemize}
\item{\textbf{Cora \cite{sen2008collective}}, \textbf{CiteSeer \cite{sen2008collective}}, \textbf{PubMed \cite{namata2012query}}.} These three datasets consist of scientific publications. Each publication are described by word vectors inferred from of predefined dictionaries. 

\item{\textbf{Wiki-CS \cite{mernyei2020wiki}.}} The Wiki-CS is a web dataset, constructed from the Wikipedia articles on computer science. The node attributes are derived from the corresponding text of the article and calculated as the average of pretrained GloVe word embeddings.

\item{\textbf{HEP-TH \cite{gehrke2003overview}.}} HEP-TH is a collection of Arxiv papers on high energy physics theory. The data covers  the period from January 1993 to April 2003, representing the complete history of HEP-TH section. To get node embeddings, we use PhysBERT \cite{10.1063/5.0238090} that is a specialized model for physics domain. 

\item{\textbf{ogbn-Arxiv \cite{yan2023comprehensive}.}} The ogbn-arxiv-TA dataset is derived from ogbn-arxiv. Each paper node’s text attributes are obtained from its title and abstract in ogbn-arxiv. For obtaining node embeddings, we utilize the pretrained language model RoBERTa \cite{liu2019roberta}. 

\item{\textbf{DBLP \cite{yan2023comprehensive}.}} DBLP is a directed graph dataset that captures citation relationships among a subset of papers from DBLP. The title and abstract of each paper are used as its node text attributes. RoBERTa model is used to produce node representations.

\item{\textbf{FB15K \cite{bollacker2008freebase}.}} FB15K is a heterogeneous network that is constructed from a knowledge base called FreeBase. Nodes in this graph are associated with the wikidata’s description as semantic information and the page views of the corresponding Wikipedia page as the node importance. FB15K includes large number of different edge types. 

\end{itemize}

\section{Link Prediction task within PINE} \label{lp_training}
For each dataset, we start with a hyperparameter optimization of GAT model to effectively solve LP problem. We adjust such parameters as number of GAT layers, hidden size, and learning rate. A grid search approach is followed. For all datasets, there is only one attention head in each layers for a transparency of a subsequent PINE score calculation. The identified hyperparameters are given in Table \ref{tab:hypers}. 

After a hyperparameter optimization step, we split all edges into train (70\%), validation (15\%), and test (15\%) groups. Training edges are not shared for message passing and supervision. There is a disjoint set of 30\% of training edges that are used to provide supervision signal during training. Validation set is used to perform early stopping. The quality of solving LP task is estimated on test set and reported in Table \ref{tab:hypers}. ROC AUC metric is calculated for the test set that is composed of positive and negative edges in 1:1 ratio. For collecting statistics on test, 5 runs are conducted. Attention weights from the first layer of GAT are taken to calculate PINE importance scores since such setting consistently produces strong results on the majority of datasets.

\begin{table}
\caption{The optimized hyperparameters of GAT network and test ROC AUC metric within LP task for different datasets for solving LP task.}
\label{tab:hypers}
\begin{tabular}{ccccc}
\toprule
Dataset & \# Layers & Hidden size & lr & AUC \\
\midrule
Cora & 1 & 512 & 0.0005 & 0.887 $\pm$ 0.001 \\
CiteSeer & 1 & 512 & 0.001 & 0.931 $\pm$ 0.002\\
PubMed & 1 & 512 & 0.001 & 0.901 $\pm$ 0.0003\\
Wiki-CS & 2 & 128 & 0.005 & 0.950 $\pm$ 0.001\\
Hep-TH & 2 & 128 & 0.0005 & 0.934 $\pm$ 0.001 \\
ogbn-Arxiv & 2 & 256 & 0.005 & 0.947 $\pm$ 0.0004 \\
DBLP & 2 & 128 & 0.01 & 0.929 $\pm$ 0.001 \\
\bottomrule
\end{tabular}
\end{table}

\section{Related work on patent analysis} \label{patent_works}

\cite{JIANG2023101402} focuses on predicting patent application outcomes by integrating text content with contextual network embeddings, showcasing the potential of combining semantic insights with structural data. Another approach \cite{zou2023eventbaseddynamicgraphrepresentation} utilizes event-based dynamic graph representation learning to forecast what types of patents companies will apply for in the next period of time to figure out their development strategies. The work \cite{patents_stock_2022} concerns the issue of the impact of core patents on the stock market. The authors of \cite{CHUNG2020120146}, \cite{LEE2018291}, and \cite{cnc_ml_tool} study the early detection of valuable patents and emerging technologies using deep and machine learning tools. 

The group of works aim to evaluate the degree of patent innovation by a sophisticated record of citations. \cite{Park2023} proposes an approach of measuring the "disruptiveness" of a patent by the extent it changes the network of citations. In turn, \cite{macher2023illusiveslumpdisruptivepatents} and \cite{petersen2023disruptionindexbiasedcitation} shed light on the deficiencies of the established disruption index. In particular, \cite{petersen2023disruptionindexbiasedcitation} studies the problem of citation inflation in the global citation graph that brings bias in disruption index evaluation. The problem with the similar purpose of identifying key contributors is called main path analysis. This direction investigates the ways of domain development and highlights the main path that includes the entities with pivotal roles. \cite{main_path_chromatography} examines the dynamics of technological development in the example of chromatography. \cite{improved_main_path} and \cite{Park_2017} suggest helpful modifications of algorithms for finding the main path.

\section{Processing of patent data} \label{patent_data}

\paragraph{Data collection.}
In this study, we use Google Patents as a source of patent documents. To obtain relevant data, we form a request to the search engine. This request constitutes a key word that concisely and clearly describes a technological domain of our interest. In the field of patenting, each patent document is associated with Cooperative Patent Classification (CPC) codes. In case we want to get more broad search results or adjust a topical coverage, we can supplement our request with CPC codes that are most relevant to the key word. After forming a request, we collect the search output of Google Patents and treat it as an initial version of a dataset. 

\paragraph{Data representation.}
A set of patents is viewed as a graph $G=(V, E)$, where vertices $v_i \in V$ stand for patents and an edge from $E$ represents citation of patent $v_i$ by patent $v_j$ and vice versa. We keep  graph undirected for simplicity. As we intend to use patent content further, we attribute each node with a concatenation of embeddings of patent title, abstract, and claims. These three sections contain condensed information on patent key points and are well-suited for its comprehensive representation. We use a language model called PatentSBERTa \cite{bekamiri2021patentsbertadeepnlpbased} that produces embeddings of size 768. As a patent is associated with the concatenation of three embeddings, the total vector dimension for node description is 2304. The relevance score of patent documents to the specified key word is measured by a cosine similarity between embeddings of the key word and a patent title. We decide to relate a key word specifically with a title because it is succinct and usually conveys the patent main idea, making relation measurement well-conditioned.

\paragraph{Prevention of redundancy in data.}
Each patent document has its own unique identification number (ID). However, a patent can be registered in different countries, be converted from an application status to the published state, or be subjected to other events that change its ID. We treat all the updated versions of a patent as one entity that represents a specific technology unit. Hence, to prevent a redundancy, we merge information from different versions of one patent, thereby maintaining a single entry in the dataset. In particular, information consolidation from patents of one meaning unit but of different IDs includes a merging process of their citation lists. One peculiarity that should be properly accounted for is that several patents may cite one patent but use its different IDs. To settle this issue, we keep all encountered IDs related to one patent in our dataset and establish the fact of citation only after inspecting all of them.

\paragraph{Data enrichment.}
For some key words, Google Patents provides incomplete search results. It means that some patents that are relevant to the key word might be missing in a search output. To tackle this problem, we perform graph enrichment procedure. We analyze the citation information of patents that are initially provided by Google Patents. The patents that cite or are cited by an original group of patents are candidates for addition to the dataset.

The enrichment process can be decomposed in steps. Firstly, we compute the median relevance scores of available set of patents toward the specified key word. Secondly, we collect a list of candidates that either cite our set or are cited by our set. Thirdly, we add a candidate to our dataset if its relevance score is higher than a current median relevance score or it is strongly connected with our set by citation relations. The last condition is that we add candidates that are from top $25\%$ by the number of connections and with relevance score that is not lower than the predefined threshold. Finally, after adding all suitable candidates, we recalculate the median relevance scores of patents in our dataset and repeat the whole procedure until we are satisfied by the size of our dataset. 

With the described enrichment process, we are able to significantly increase the number of nodes in the graph. As we add patents that meet one of two conditions on relevance and connectedness. Following these conditions, we force the distribution of relevance scores to move towards higher values and allow to enrich graph only with structurally highly involved patents. 

\paragraph{Structural reformation of data.}
After data collection and citation establishment, we can build a citation graph. Usually, such graphs have several connected components. The components can be sorted in order of size of subgraphs they represent. The largest connected component we call the main one. We assume that the majority of the influential and relevant patents are present in it. The smaller connected components tend to contain patent that stand aside from the topic covered by patents from the main one, thereby less relevant from structural and topical points of view. For further consideration, we leave only patents forming the main component.

\section{Metrics for supervised performance validation} \label{supervised_metrics}

$\mathrm{\textbf{NDCG@k}}$ is a measure for assessing the ranking quality of the ordered list of elements. To calculate it, we sort the nodes by the predicted importance and denote the corresponding ground truth score at the position $i$ as $\mathrm{imp}_i$. The value of $\mathrm{DCG@k}$ is calculated as:
\begin{equation}
\mathrm{DCG@k} = \sum_{i=1}^{k} \frac{\mathrm{imp_i}}{\log_2{(i+1)}}.
\end{equation}

To compute ideal $\mathrm{DCG@k}$, designated as $\mathrm{IDCG@k}$, nodes are sorted by their ground truth scores, after which $\mathrm{DCG@k}$ is evaluated for the resulting list. Finally, $\mathrm{NDCG@k}$ is defined as:
\begin{equation}
\mathrm{NDCG@k} = \frac{\mathrm{DCG@k}}{\mathrm{IDCG@k}}.
\end{equation}

$\mathrm{\textbf{Spearman}}$ estimates a correlation between the ground truth and predicted importance scores after converting into ranks, denoted by $r$ and $r^{\prime}$. The $\mathrm{Spearman}$ correlation value is calculated by:

\begin{equation}
\mathrm{Spearman} = \frac{\sum_i(r_i - \overline{r})(r^{\prime}_i - \overline{r^{\prime}})}{\sqrt{\sum_i(r_i - \overline{r})^2 \vphantom{\sum_i(r^{\prime}_i - \overline{r^{\prime}})^2}}\sqrt{\sum_i(r^{\prime}_i - \overline{r^{\prime}})^2}},
\end{equation}

where $\overline{r}$ and $\overline{r^{\prime}}$ correspond to the mean values over all ground truth and predicted ranks, respectively.

$\mathrm{\textbf{Precision@k}}$ reflects the overlap between top-$k$ sets of nodes obtained with ground truth and predicted importance scores. We denote them as $S^{gt}_{top\text{-}k}$ and $S^{pred}_{top\text{-}k}$. Then, $\mathrm{Precision@k}$ score is determined as:

\begin{equation}
\mathrm{Precision@k} = \frac{|S^{gt}_{top\text{-}k} \cap S^{pred}_{top\text{-}k}|}{k}
\end{equation}

\section{Computational time measurements} \label{computation_time}

\begin{figure}
\centering
\includegraphics[width=\linewidth]{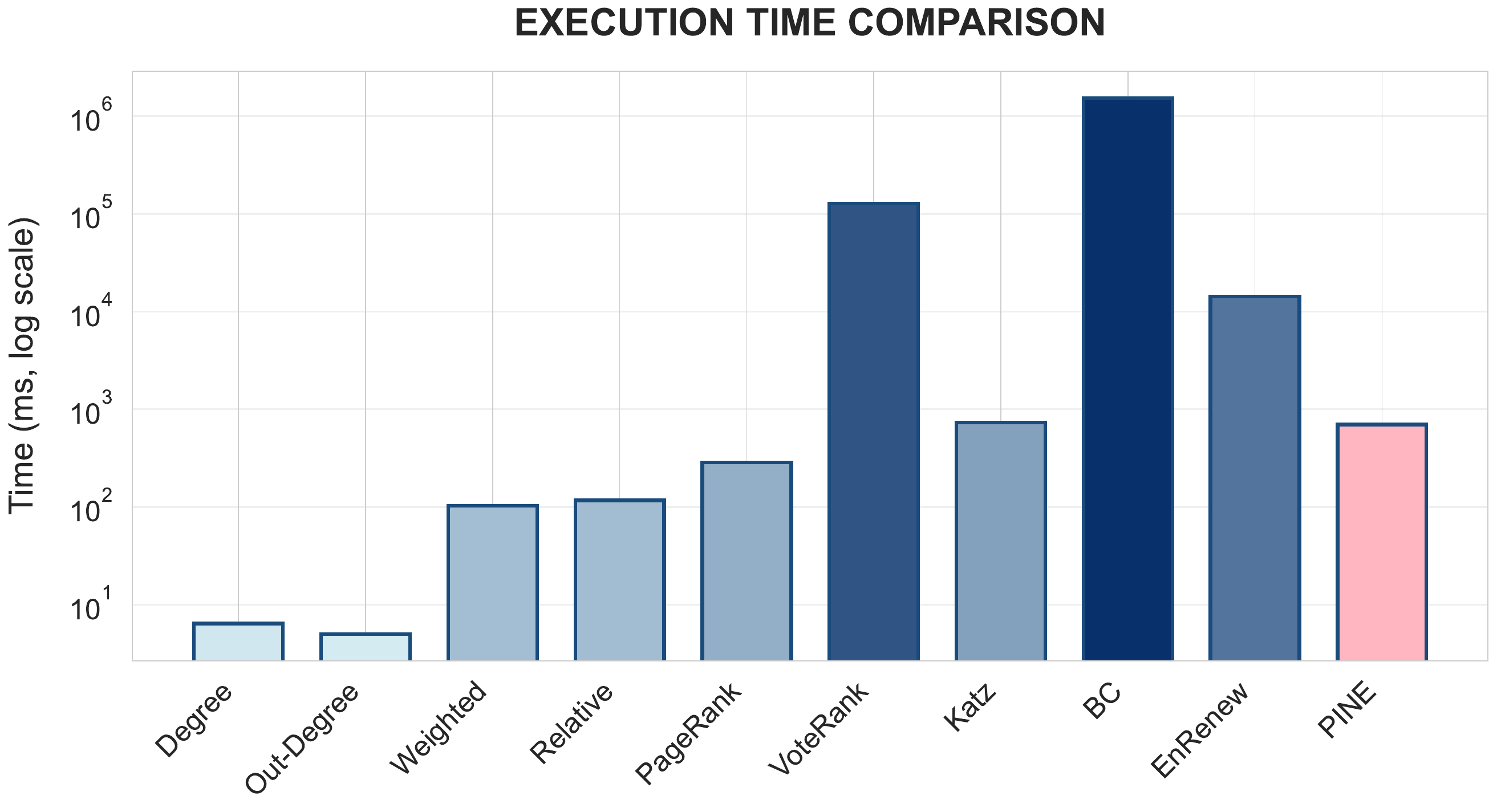}
\caption{Running time comparison on PubMed dataset. PINE takes a reasonable time to infer node importance scores. Although, a quality of the produced sets of important nodes are much higher.}
\label{fig:time}
\end{figure}

To compare PINE with other existing approaches from an efficiency prospective, we estimate its running time. For PINE, running time is a sum of time for training and inference stages. The results of time measurements on PubMed datasets are shown in Figure \ref{fig:time}. As we can see, PINE needs a reasonable time to get importance scores for each node. A part of approaches, namely Betweenness Centrality and VoteRank, face computations problems on large graphs.

\end{document}